\documentclass[final]{l4dc2024}

\usepackage[subtle]{savetrees}
\usepackage{lineno,hyperref}
\modulolinenumbers[5]

\usepackage{tikz}
\usepackage{comment}
\usepackage{pgfplots}
\usepackage{slashbox}
\usepackage{caption}
\usepackage{xcolor}
\allowdisplaybreaks  
\usepackage{mathtools}
\usepackage{hyperref}
\usepackage{cleveref}
\usepackage{multirow}
\usepackage{optidef}
\usepackage{todonotes}
\usepackage[frozencache,cachedir=minted-output]{minted}

\pgfplotsset{compat=1.18}

\newtheorem{problem}{Problem}

\title[Multi-Modal Conformal Prediction Regions]{Multi-Modal Conformal Prediction Regions with Simple Structures by Optimizing Convex Shape Templates}
\usepackage{times}
\author{%
 \Name{Renukanandan Tumu}$^{*}$$^{1}$ \Email{nandant@seas.upenn.edu} \AND
\Name{Matthew Cleaveland}$^{*}$$^{1}$ \Email{mcleav@seas.upenn.edu}\AND
 \Name{George J. Pappas}$^{1}$ 
 \Email{pappasg@seas.upenn.edu}\AND
 \Name{Rahul Mangharam}$^{1}$ 
 \Email{rahulm@seas.upenn.edu}\AND
  \Name{Lars Lindemann}$^{2}$ 
 \Email{llindema@usc.edu}\AND
 \addr $^{*}$ Indicates equal contribution\\
 \addr $^{1}$ Department of Electrical \& Systems Engineering, University of Pennsylvania\\
\addr $^{2}$ Thomas Lord Department of Computer Science, University of Southern California  
}

\begin{document}

\maketitle

\begin{abstract}
    Conformal prediction is a statistical tool for producing prediction regions for machine learning models that are valid with high probability. A key component of conformal prediction algorithms is a \emph{non-conformity score function} that quantifies how different a model's prediction is from the unknown ground truth value. Essentially, these functions determine the shape and the size of the conformal prediction regions. While prior work has gone into creating score functions that produce multi-model prediction regions, such regions are generally too complex for use in downstream planning and control problems. We propose a method that optimizes parameterized \emph{shape template functions} over calibration data, which results in non-conformity score functions that produce prediction regions with minimum volume. Our approach results in prediction regions that are \emph{multi-modal}, so they can properly capture residuals of distributions that have multiple modes, and \emph{practical}, so each region is convex and can be easily incorporated into downstream tasks, such as a motion planner using conformal prediction regions. Our method applies to general supervised learning tasks, while we illustrate its use in time-series prediction. We provide a toolbox and present illustrative case studies of F16 fighter jets and autonomous vehicles, showing an up to 68\% reduction in prediction region area compared to a circular baseline region. 
\end{abstract}

\section{Introduction}
\label{sec:intro}
\begin{figure}[t]
    \centering
    \includegraphics[width=0.8\linewidth]{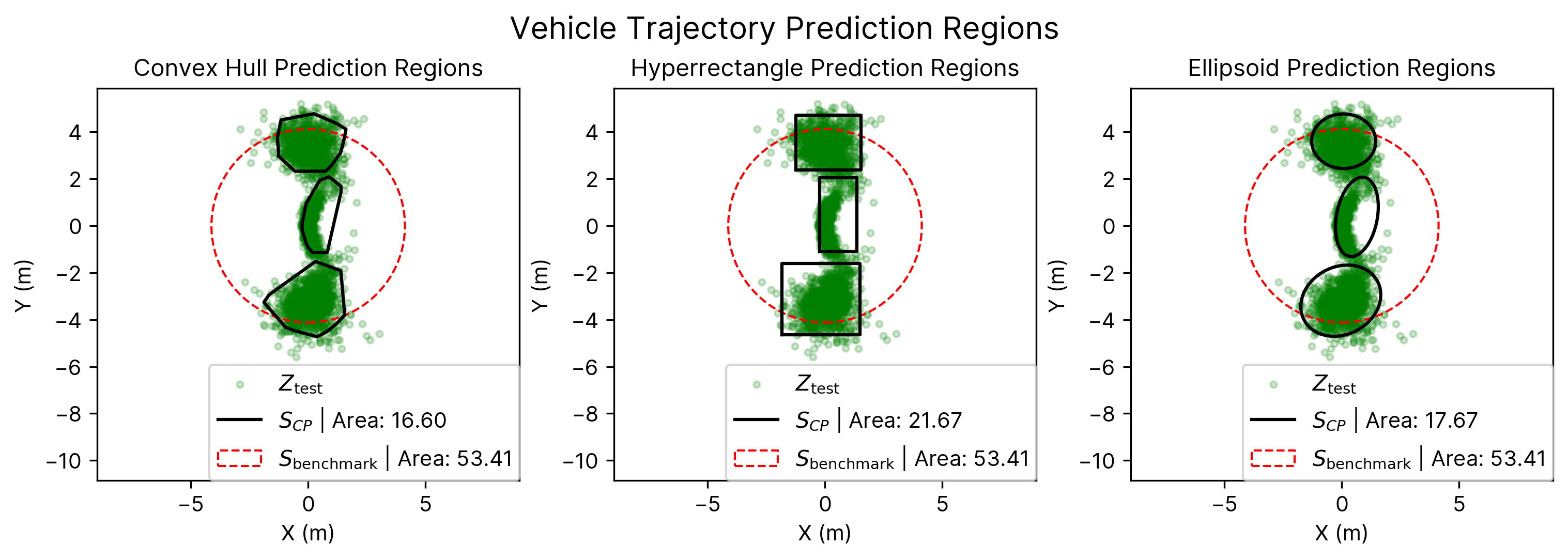}
    \caption{Vehicle Trajectory Prediction regions, $S_{CP}$, plotted alongside benchmark prediction regions $S_{\text{benchmark}}$, which are based on the 2-norm of the residual. All methods achieve the target $90\%$ coverage rate. The Convex Hull, Hyperrectangle, and Ellipsoid Regions are $68.92\%$, $59.43\%$, and $66.92\%$ smaller respectively.}
    \label{fig:region-comparison}
\end{figure}

Conformal prediction (CP) has emerged as a popular method for statistical uncertainty quantification \cite{shafer2008tutorial,vovk2005algorithmic}. It aims to construct regions around a predictor's output, called prediction regions, that contain the true but unknown quantity of interest with a user-defined probability. CP only requires relatively weak assumptions of the data or the predictor itself, and instead one only needs a calibration dataset. This means that CP can be applied to learning-enabled predictors, such as neural networks \cite{angelopoulos2021gentle}. 


Conformal prediction regions often take the form $\{ y | R(y,\hat{y}) \leq C \}$, where $\hat{y}$ is a prediction and $R(y,\hat{y}) \in \mathbb{R}$ is a \emph{non-conformity score function}. This function quantifies the difference between the ground truth $y$ and the prediction $\hat{y}$, while $C \in \mathbb{R}$ is a bound produced by the CP procedure. The choice of non-conformity score function plays a vital role as it defines what shape and size the prediction regions take. For example, using the L2 norm on the error between $y$ and $\hat{y}$ ensures that the CP regions will be circles or hyperspheres depending on the dimension of $y$. However, if the distribution of errors from the predictor does not resemble a sphere, e.g., when there are dependencies across dimensions, then the L2 norm is not the right choice as it will result in unnecessarily large prediction regions. While there is initial work towards the design of non-conformity scores for robotic planning and control, see e.g., \cite{tumu2023physics} and \cite{cleaveland2023conformal}, a systematic approach to generate non-conservative conformal prediction regions for these applications is missing. Additionally, using the L2 norm as a non-conformity score function does not allow for disjoint prediction regions. This further leads to overly large prediction regions when prediction errors have multi-modal distributions, such as when predicting which way a vehicle will turn at an intersection, as shown by the dotted red circles in~\Cref{fig:region-comparison}. To address this issue, \cite{zecchin2023forking} and \cite{wang_probabilistic_2022} build prediction regions using multiple predictions. \cite{feldman_calibrated_2022} generate prediction regions in a latent space and then transform the region into the original domain to get non-convex regions. \cite{lei2011efficient,Smith2014ConformalKDE} employ kernel density estimators (KDEs) which can capture disjoint prediction regions. However, density based prediction regions are mathematically difficult to handle and not suitable for real-time decision making. \cite{Izbicki2022, han2022split} use conditional probability predictors to generate efficient prediction regions, but these shapes can be too complex to use in downstream tasks. For example, \cite{lindemann2022safe,dixit2023adaptive} use conformal prediction regions for model predictive control  which cannot handle prediction regions from KDEs efficiently. With this in mind, we also seek to produce conformal prediction regions that are \textit{practical}, in the sense that they have simple convex structure. In pursuit of practicality, we will optimize over template shapes under suitable optimality criteria.

\noindent \textbf{Contributions. } To address the conservatism from improper choices of non-conformity score functions, this paper proposes using optimization to create non-conformity score functions that \emph{produce non-conservative conformal prediction regions that are  multi-modal and practical}. Our main idea is to use an extra calibration dataset to i) cluster the residuals of this calibration data to identify different modes in the error distribution, ii) define parameterized shape generating functions which specify template shapes, iii) solve an optimization problem to fit parameterized shape functions for each cluster over the calibration data while minimizing the volume of the shape template, and iv) use the resulting set of shape template functions to define a non-conformity score function. Finally, we use a separate calibration dataset to apply CP using the new non-conformity score function. Our contributions are as follows:
\begin{itemize}
    \item We propose a  framework for generating non-conformity score functions that result in non-conservative conformal prediction regions that are multi-modal and practical for downstream tasks. We capture multi-modality using clustering algorithms and obtain non-conservative convex regions by fitting parameterized shape template functions to each cluster. 
    \item We provide a python toolbox of our method that can readily be used. We further demonstrate that our method produces non-conservative conformal prediction regions on case studies of F16 fighter jets and autonomous vehicles, showing an up to 68\% reduction in prediction region area compared to an L2 norm region.
\end{itemize}

\noindent \textbf{Related Work.} The original conformal prediction approach, introduced by \cite{vovk2005algorithmic,shafer2008tutorial} to quantify uncertainty of prediction models, required training one prediction model per training datapoint, which is computationally intractable for complex predictors. To alleviate this issue, \cite{papadopoulos2008inductive} introduce inductive conformal prediction, which can also be referred to as split conformal prediction. This method employs a calibration dataset for applying conformal prediction. Split conformal prediction has been extended to allow for quantile regression \cite{romano2019conformalized}, to provide conditional statistical guarantees \cite{vovk2012conditional}, and to handle distribution shifts \cite{tibshirani2019conformal,fannjiang2022conformal}. Applications of split conformal prediction include out-of-distribution detection \cite{Kaur2022,kaur2023codit}, guaranteeing safety in autonomous systems \cite{luo2022sample}, performing reachability analysis and system verification \cite{hashemi2023data,lindemann2022conformal}, and bounding errors in F1/10 car predictions \cite{tumu2023physics}. Additionally, prior works have constructed probably approximately correct prediction sets around conformal prediction regions \cite{vovk2012conditional,angelopoulos2022conformal}.

However, the aforementioned methods use non-conformity score functions that may not fit the residuals of their predictors well, which could result in unnecessarily large prediction regions. Previous works have addressed this limitation by employing density estimators as non-conformity score functions. In \cite{lei2011efficient,lei2013distribution,lei2014distribution,Smith2014ConformalKDE} the authors use kernel density estimators (KDEs) to produce conformal prediction regions. Another work uses conditional density estimators, which estimate the conditional distribution of the data $p(Y|X)$, where $Y$ is the predicted variable, and $X$ is the input variable, to produce non-conformity scores \cite{Izbicki2022}. \cite{han2022split} partitions the input space and employs KDEs over the partitions to compute density estimates, which allow for conditional coverage guarantees. In \cite{stutz2022ICLR}, the authors encode the width of the generated prediction sets directly into the loss function of a neural network while training. While these approaches can produce small prediction regions, they may not have analytical forms that are easy for downstream tasks to make use of. In a different vein, \cite{bai2022efficient} uses parameterized conformal prediction sets and expected risk minimization to produce small prediction regions. 
Some work has also gone into producing multi-modal prediction regions. These works use set based predictors to compute multiple predictions and conformalize around these sets of predictions \cite{wang2022probabilistic,zecchin2023forking,parente2023conformalized}. Our work on the other hand, does not require a set of predictions to generate multi-modal regions.

\section{Preliminaries: Conformal Prediction Regions}
\label{sec:preliminaries}

\textbf{Split Conformal Prediction.} Conformal prediction was introduced in \cite{vovk2005algorithmic,shafer2008tutorial} to obtain valid prediction regions, e.g., for complex predictive models such as neural networks. Split conformal prediction, proposed in \cite{papadopoulos2008inductive}, is a computationally tractable variant of conformal prediction where  a calibration dataset is available that has not been used to train the predictor. Let $R_0,R_1\hdots,R_{n}$ be $n+1$ exchangeable random variables\footnote{Exchangeability is a weaker assumption than being independent and identically distributed (i.i.d.).}, usually referred to as the \emph{nonconformity scores}. Here, $R_0$ can be viewed as a test datapoint, and $R_i$ with $i\in\{1,\hdots,n\}$ as a set of calibration data. The nonconformity scores are often defined as $R:=\|Y-h(X)\|$ and $R_{i}:=\|Y_{i}-h(X_{i})\|$ where $h$ is a predictor that attempts to predict the output from the input. Our goal is now to obtain a probabilistic bound for $R_0$ based on $R_{1},\hdots,R_{n}$. Formally, given a failure probability $\delta\in (0,1)$, we want to compute a constant $C$ so that\footnote{More formally, we would have to write $C(R_{1},\hdots,R_{n})$ as the prediction region $C$ is a function of $R_{1},\hdots,R_{n}$, e.g., the probability measure Prob$(\cdot)$ is defined over the product measure of $R_0,R_1\hdots,R_{n}$.}
\begin{align}\label{eq:cpGuaranteeVanilla}
    \text{Prob}(R_0\le C)\ge 1-\delta.
\end{align}
In conformal prediction, we compute $C:=\text{Quantile}(\{ R_{1}, \hdots, R_{n}, \infty \},1-\delta)$ which is the $(1-\delta)$th quantile of the empirical distribution of the values $R_{1},\hdots,R_{n}$ and $\infty$. Alternatively, by assuming that $R_{1},\hdots,R_{n}$ are sorted in non-decreasing order and by adding $R_{n+1}:=\infty$, we can obtain $C=R_{p}$ where $p:=\lceil (n+1)(1-\delta)\rceil$ with $\lceil \cdot\rceil$ being the ceiling function, i.e., $C$ is the $p$th smallest nonconformity score. By a quantile argument, see \cite[Lemma 1]{tibshirani2019conformal}, one can prove that this choice of $C$ satisfies ~\Cref{eq:cpGuaranteeVanilla}. Note that $n\ge \lceil (n+1)(1-\delta)\rceil$ is required to hold to obtain meaningful, i.e., bounded, prediction regions.


\noindent \textbf{Existing Choices for Non-Conformity Score Functions.} The guarantees from \eqref{eq:cpGuaranteeVanilla} bound the non-conformity scores, and we need to convert this bound into prediction regions. Specifically,  let $(X,Y)$ and $(X_i,Y_i)$ with $i\in\{1,\hdots,n\}$ be test and calibration data, respectively, drawn from a distribution $\mathcal{D}$, with $X,X_i \in \mathcal{X} \subseteq \mathbb{R}^{l}$ and $Y,Y_i \in \mathcal{Y} \subseteq \mathbb{R}^{p}$. Assume also that we are given a predictor $h: \mathcal{X} \rightarrow \mathcal{Y}$. First, we define a \emph{non-conformity score function} $R$, which maps outputs and predicted outputs to the non-conformity scores from the previous section as $R: \mathcal{Y} \times \mathcal{Y} \rightarrow \mathbb{R}$.

Then, for a non-conformity score $R(Y,\hat{Y})$ with prediction $\hat{Y}:=h(X)$ and a constant $C$ that satisfies \eqref{eq:cpGuaranteeVanilla}, e.g., obtained from calibration data $R(Y_i,\hat{Y}_i)$ with $\hat{Y}_i:=h(X_i)$ using conformal prediction, we define the prediction region $S_{CP}$ as the set of values in $\mathcal{Y}$ that result in a non-conformity score not greater than $C$, i.e., such that
\begin{equation}
    S_{CP} := \{ y \in\mathcal{Y} | R(y,\hat{Y}) \leq C \}.
\end{equation}

The choice of the score function $R$ greatly affects the shape and size of the prediction region $S_{CP}$. For example, if we use the L2 norm as $R(Y,\hat{Y}):=\|Y-\hat{Y}\|_2$, then the conformal prediction regions will be hyper-spheres (circles in two dimensions). However, the errors of the predictor $h$ may have asymmetric, e.g., more accurate in certain dimensions, and multi-modal distributions, which will result in unnecessarily conservative prediction regions. We are interested in constructing non-conservative and multi-modal prediction regions, e.g., as shown in~\Cref{fig:region-comparison}. 

Often non-conformity score functions are fixed a-priori, e.g., as the aforementioned L2 norm distance \cite{lindemann2022safe} or by using softmax functions for classification tasks \cite{angelopoulos2021gentle}. More tailored functions were presented in \cite{tumu2023physics} for F1/10 racing applications, in  \cite{kaur2022codit} for predictor equivariance, and in  \cite{cleaveland2023conformal} for multi-step prediction regions of time series.


Data-driven techniques instead compute non-conformity scores from data. Existing techniques generally rely on density estimation techniques  which aim to estimate the conditional distribution $p(Y|X)$, see \cite{lei2011efficient}. Let $\hat{p}(Y|X)$ denote an estimate of $p(Y|X)$. One can then use the estimate $\hat{p}(Y|X)$ to define the non-conformity score as $R(Y,X) := -\hat{p}(Y|X)$. For this non-conformity score function, one can apply the conformal prediction to get a bound $C$ which results in the prediction region $S_{CP} = \{ y | \hat{p}(y|X) \leq C \}$. These regions can take any shape and potentially be multi-modal. However, these regions are difficult to work with in downstream decision making tasks, especially if the model used to form $\hat{p}(Y|X)$ is complex (e.g. a deep neural network), as $S_{CP}$ can be difficult to recover.





\noindent \textbf{Problem Formulation.} In this work, we present a combination of a data-driven technique with parameterized template non-conformity score functions.  As a result, we obtain \emph{parameterized conformal prediction regions} which we  denote as $S_{CP,\theta}:=\{ y \in\mathcal{Y} | R_\theta(y,\hat{Y}) \leq C \}$ where $\theta$ is a set of parameters. Our high level problem is now to find values for $\theta$ that minimize the size of $S_{CP,\theta}$ while still achieving the desired coverage level $1-\delta$.
\begin{problem}\label{prob}
    Let $(X,Y) \sim \mathcal{D}$ be a random variable, $D_{cal}:=\{ (X_{1},Y_{1}), \hdots, (X_{n},Y_{n}) \}$ be a calibration set of random variables exchangeably drawn from $\mathcal{D}$,  $h: \mathcal{X} \rightarrow \mathcal{Y}$ be a predictor, and  $\delta\in(0,1)$ be a failure probability. Define parameterized template non-conformity score functions $R_\theta(Y,h(X))$ for parameters $\theta$ that result in convex multi-modal prediction regions $S_{CP,\theta}$, and use the calibration set $D_{cal}$ to solve the  optimization problem:
    \begin{subequations}
    \begin{align}
    \min_{\theta} \;\; &\text{Volume}(S_{CP,\theta})\label{eq:gen_optimization_problem}\\
    \text{s.t.} \;\; & \text{Prob}\left( Y \in S_{CP,\theta} \right) \geq 1-\delta
    \end{align}
    \end{subequations}
\end{problem}

We cannot exactly solve \eqref{eq:gen_optimization_problem}, especially since one can run into trouble when using the same data to both compute the values of $\theta$ and compute the prediction region $S_{CP,\theta}$ based on $\theta$. So we aim to find heuristic solutions to \eqref{eq:gen_optimization_problem}, with an added focus on having $S_{CP,\theta}$ be multi-modal and practical for downstream tasks.

\section{Computing Convex Multi-Modal Conformal Prediction Regions}
\label{sec:solution-formulation}

To enable multi-modal prediction regions, we first cluster the residuals $Y-\hat{Y}$ over a subset $D_{cal,1}$ of our calibration data $D_{cal}$, i.e., $D_{cal,1}\subset D_{cal}$. More specifically, we perform a \textbf{density estimation} step by using Kernel Density Estimation (KDE) to find high-density modes of residuals in $D_{cal,1}$. We then perform a \textbf{clustering} step by using Mean Shift Clustering to identify multi-modality in the high-density modes of the KDE. We next perform a \textbf{shape construction} step by defining parameterized \emph{shape template functions} and by fitting a separate shape template function to each cluster. These shape template functions generate convex approximations of the identified clusters. We then perform a \textbf{conformal prediction} step where we combine all shape template functions into a single non-conformity score function. Finally, we apply conformal prediction to this non-conformity score over the calibration data $D_{cal,2}:=D_{cal}\setminus D_{cal,1}$. The use of a separate calibration set $D_{cal,2}$ guarantees the validity of our method. We explain each of these steps now in detail.

\paragraph{Density Estimation} Let us define the residuals $Z_i := Y_i - \hat{Y}_i$ with $\hat{Y}_i:=h(X_i)$ for each calibration point $(X_i,Y_i)\in D_{cal,1}$. We then define the set of residuals $Z := \{Z_1, \ldots, Z_{n_1}\}$ where $n_1:=|D_{cal,1}|$. We seek to understand the distribution of these residuals to build multi-modal prediction regions. For this purpose, we perform Kernel Density Estimation (KDE) over the residuals $Z$ of $D_{cal,1}$. Note that we can use any other density estimation method here. In doing so, we will be able to capture high-density modes of the residual distribution.

KDE is an approach for estimating the probability density function of a variable from data \cite{parzen_estimation_1962, rosenblatt_remarks_1956}. The estimated density function using KDE takes the form


\begin{equation}\label{eq:KDEeqn}
    \hat{p}(z | \bar{K},b,Z_1,\ldots,Z_{n_1}) = \frac{1}{n_1b} \sum_{i=1}^{n_1} \bar{K}\left( \frac{z-Z_i}{b} \right)
\end{equation}
where $Z_1,\hdots,Z_{n_1}$ are the residuals from $D_{cal,1}$, $\bar{K}$ is a kernel function, and $b$ is the bandwidth parameter. The kernel $\bar{K}$ must be a non-negative, real-valued function. In this work, we use the Gaussian kernel $\bar{K}(z) := \frac{1}{\sqrt{2\pi}}\exp(-z^2/2)$, which is the density of the standard normal distribution. The bandwidth parameter $b$ controls how much the density estimates spread out from each residual $Z_i$, with larger values causing the density estimates to spread out less. We use Silverman's rule of thumb \cite{silverman_density_1986} to select the value of $b$. Using a combination of KDE with Silverman's rule of thumb yields a parameter-free method of estimating the probability density of a given variable.



We use the KDE $\hat{p}$ to find a set $\bar{L} \subseteq \mathcal{Y}$ that covers a $1-\delta$ portion of the residuals, i.e., we want to compute  $\bar{L}$ such that $1-\delta \leq \int_{\bar{L}} \hat{p}(z | \bar{K},b,Z_1,\ldots,Z_{n_1})\, dz$. As $\bar{L}$ is difficult to compute in practice, our algorithm first grids the $Z$ domain. The density of the grid can be set based on the density of the data, and is a key driver of the runtime of the algorithm. A high density grid can result in smoother, smaller regions, at the expense of memory and runtime. This gridding approach can be expensive when $Z$ is of high dimension. Let $J$ be the number of grid cells and let $g_j\subseteq \mathcal{Y}$ denote the $j$th grid cell.  Next, we compute $\hat{p}(z_j^c | \bar{K},b,Z_1,\ldots,Z_{n_1})$ for a single point $z_j^c\in g_j$ of each grid cell (e.g., its center) and multiply $\hat{p}(z_j^c | \bar{K},b,Z_1,\ldots,Z_{n_1})$ by the volume of the grid cell to obtain its probability density. Finally, we sort the probability densities of all grid cells in decreasing order and add grid cells (start from high-density cells) to $\bar{L}$ until the cumulative sum of probability densities in $\bar{L}$ is greater than $1-\delta$. Having computed high density modes in $\bar{L}$, we construct the discrete set $L:=\{z_1^c,\hdots,z_J^c\}$. We note that we will get valid prediction regions despite this discretization.

\paragraph{Clustering \label{sec:clustering}}  In the next step, we identify clusters of points within $L$ toward obtaining multi-modal prediction regions. To accomplish this, we use the Mean Shift algorithm \cite{comaniciu_mean_2002} since it does not require a pre-specified number of clusters. The algorithm attempts to find local maxima of the probability density $\hat{p}$ within $L$. The algorithm requires a single bandwidth parameter, which we estimate from data using the bandwidth estimator package in \cite{scikit-learn}. Due to space limitations, we direct the reader to \cite{comaniciu_mean_2002} for more details. Once the local maxima are found, we group all of the points within $L$ according to their nearest maxima, resulting in the set $L =: \{L_1, \ldots, L_K\}$, where $K$ denotes the number of clusters. 



\paragraph{Shape Construction \label{sec:shapeConstruction}} For each cluster $L_k\in L$, we now construct convex over-approximations. Our approximations are defined by parameterized \emph{shape template functions} $f_{\theta_k}$ and take the form \(S_{\theta_k} = \{z|f_{\theta_k}(z) \leq 0\}\). We specifically consider shape template functions for ellipsoid, convex hulls, and hyperrectangles (details are provided below). 
Given a cluster of points $L_k$ and a parameterized template function $f_{\theta_k}$, we find the parameters ${\theta_k}$ that minimize the volume of $S_{\theta_k}$ while covering all of the points in $L_k$. This is formulated as the following optimization problem:
\begin{subequations}\label{eq:general_shape_optim}
    \begin{align}
    \min_{\theta_k} \;\; &\textrm{Volume}(S_{\theta_k})\\
    \text{s.t.} \;\; &S_{\theta_k}= \{ z | f_{\theta_k}(z) \leq 0 \}\\
    \;\; &z\in S_{\theta_k} \;\; \forall z \in L_k.
\end{align}
\end{subequations}
After solving this optimization problem for each cluster $L_k\in L$, we get the set of shapes $S_c := \{S_{\theta_1}, \ldots, S_{\theta_k}\}$. Below, we provide our three choices of template shapes. The choice of shape template is a hyperparameter of our algorithm. 


\emph{Ellipsoid: } The definition for an ellipsoid in $\mathbb{R}^p$ parameterized by ${\theta_k}:=\{ Q \succ 0 \in \mathbb{R}^{p\times p}, c \in \mathbb{R}^p\}$ is $\text{Ell}_{\theta_k} \coloneqq \{ z \in \mathbb{R}^p | (z-c)^T Q (z-c) \leq 1 \}$. The shape template function for an ellipsoid is 




\begin{equation}
     f_{\theta_k}(z):= (z-c)^T Q (z-c) - 1.
\end{equation}
We solve the problem in \Cref{eq:general_shape_optim} for $\theta_k$ (consisting of $Q$ and $c$)  under this parameterization  by using CMA-ES, a genetic algorithm \cite{hansen_reducing_2003,hansen_cma-espycma_2023}. 


\emph{Convex Hull:} The definition for a Convex Hull in $\mathbb{R}^p$ parameterized by ${\theta_k} := \{A \in \mathbb{R}^{r\times p}, b\in \mathbb{R}^r\}$ is $\textrm{CXH}_{\theta_k} := \left\{z \in \mathbb{R}^p | Az - b \leq 0 \right\}$, where $r$ is the number of facets in the Convex Hull. This way, the shape template function for a convex hull is
\begin{equation}
    f_{\theta_k}(z) := \max_{j\in \{1,\ldots, r\}} A_{j} z -b_j
\end{equation}
where $A_j$ and $b_j$ denote the $j^{th}$ row of $A$ and $b$, respectively. We solve the problem in \Cref{eq:general_shape_optim} for $\theta_k$ (consisting of $A$ and $b$)  under this parameterization by using the Quickhull Algorithm from \cite{barber_quickhull_1996}. This algorithm generates a convex polytope that contains every point in $L_k$.


\emph{Hyper-Rectangle: } The definition for a (non-rotated) hyper-rectangle parameterized by ${\theta_k} := \{b_{min} \in \mathbb{R}^p, b_{max} \in \mathbb{R}^p\}$ is $\textrm{HypRect}_{\theta_k} := \left\{ z\in \mathbb{R}^p | b_{min} \leq z \leq b_{max} \right\}$. Consequently, the shape template function for a hyper-rectangle is




\begin{equation}
    f_{\theta_k}(z) := \max_{j\in \{1,\ldots, p\}} \max \left\{ b_{min,j} - z_j, z_j - b_{max,j} \right\}
\end{equation}
We solve the problem in \Cref{eq:general_shape_optim} for $\theta_k$ (consisting of $b_{min}$ and $b_{max}$)  under this parameterization by computing the element-wise minimum and maximum of the datapoints in $L_k$.

\paragraph{Conformalization} Note that the set $S_c=\{S_{\theta_1}, \ldots, S_{\theta_k}\}$, while capturing information about the underlying distribution of residuals, may not be a valid prediction region. To obtain valid prediction regions, we define a new nonconformity score based on the shape template functions $\{f_{\theta_1},\hdots,f_{\theta_k}\}$ to which we then apply conformal prediction over the second dataset $D_{cal,2}$. To account for scaling differences in $f_{\theta_k}$, which each describe different regions, we normalize first. Specifically, we compute a normalization constant $\alpha_k$ for each $f_{\theta_k}$ as
\begin{gather}
    \mathcal{R}_k :=\left\{ f_{\theta_k}(z) \middle| z \in D_{cal,1} \right\} \notag\\
    \alpha_k := 1/\left(\text{Quantile}(\mathcal{R}_k, 1-\delta) - \min(\mathcal{R}_k)\right)
    \label{eq:normalizing_constant}
\end{gather}
We then define the non-conformity score for each shape as the normalized shape template function
\begin{equation}
    R_{\theta_k}(z) := \alpha_k f_{\theta_k}(z).
\end{equation}
Finally, we define the joint non-conformity score over all shapes using the smallest normalized non-conformity score as 
\begin{equation} \label{eq:nonconformFuncShapes}
    R_{S_c}(z) := \text{min}(R_{\theta_1}(z),\hdots,R_{\theta_K}(z))
\end{equation}
We remark here that we take the minimum because we only need the residual point $z$ to lie within one shape. We can then apply conformal prediction to this non-conformity score function over the second dataset $D_{cal,2}$ to obtain a valid multi-modal prediction region. The next result follows immediately by \cite[Lemma 1]{tibshirani2019conformal} and since we split $D_{cal}$ into $D_{cal,1}$ and $D_{cal,2}$.


\begin{theorem}\label{thm:cp}
Let the conditions from Problem \ref{prob} hold. Let $R_{S_c}$ be the non-conformity score function according to equation \eqref{eq:nonconformFuncShapes} where the parameters $\theta_1,\hdots,\theta_K$ are obtained by solving \Cref{eq:general_shape_optim}. Define $R:=R_{S_c}(Y-\hat{Y})$ for the random variable $(X,Y)\sim \mathcal{D}$ and $R_{i}:=R_{S_c}(Y_{i}-\hat{Y}_{i})$ for the calibration data $(X_i,Y_i)\in D_{cal,2}$ with $i\in\{n_1+1,\hdots, n\}$. Then, it holds that
\begin{equation}\label{eq:cpGuarantee}
    Prob(R\leq C) \geq 1-\delta
\end{equation}
where $C\coloneqq Quantile(\{ R_{n_1+1},\hdots,R_{n},\infty \}, 1-\delta)$.
\end{theorem}
To convert the probabilistic guarantee in equation \eqref{eq:cpGuarantee} into valid prediction regions, we note that 
\begin{equation}
    R = \text{min}(R_{\theta_1}(Z),\hdots,R_{\theta_K}(Z)) \leq C \iff \exists k\in \{ 1,\hdots,K \} \; s.t. \; f_{\theta_k}(Z) \leq C/\alpha_k.
\end{equation}
For a prediction $\hat{Y}$, this means in essence that a valid prediction region is defined by
\begin{equation}
    S_{CP} := \{ y | \exists k \in \{ 1, \hdots, K \}  \; s.t. \; f_{\theta_k}(y-\hat{Y}) \leq C/\alpha_k \} 
 = \cup_{k=1}^{K} \{ y | f_{\theta_k}(y-\hat{Y}) \leq C/\alpha_k \}
\end{equation}
Intuitively, the conformal prediction region $S_{CP}$ is the union of the prediction regions around each shape in $S_c$ which illustrates its multi-modality. We summarize our results as a Corollary. 

\begin{corollary}
    Let the conditions of \Cref{thm:cp} hold. Then, it holds that $\text{Prob}\left( Y \in S_{CP} \right) \geq 1-\delta$.
\end{corollary}

\paragraph{Dealing with Time-Series Data} Let us now illustrate how we can handle time series data. Assume that $P_0,P_1,\hdots, P_T\in \mathbb{R}^{p(T+1)}$ is a time series of length $T$ that follows the distribution $\mathcal{D}$. At time $t>0$, we observe the inputs $X:=(P_0,\hdots,P_t)$ and want to predict the outputs $Y:=(P_{t+1},\hdots,P_T)$ with a trajectory predictor $h$, e.g., a recurrent neural network. Our calibration dataset $D_{cal,1}$ consists of pairs $(X,Y)\sim\mathcal{D}$ where $X=(P_0,\hdots,P_t)$ and $Y=(P_{t+1},\hdots,P_{T})$ and the residuals are $Z_\tau = P_{\tau}-h(X)_{\tau}$ for $\tau = t+1,\hdots, T$. Now, our desired prediction region should contain every future value of the time series, $P_{t+1},\hdots,P_{T}$, with probability $1-\delta$. To achieve this in a computationally efficient manner, we follow the previously proposed optimization procedure for each future time $\tau\in \{t+1,\hdots, T\}$ independently again with a desired coverage of $1-\delta$. As a result, we get a non-conformity score $R_{S_c}^\tau$ for each time $\tau$, similarly to \Cref{eq:nonconformFuncShapes}. We normalize these scores over the future times, obtaining normalization constants $\beta_\tau$, as in \Cref{eq:time_normalizing_constant}. Finally, we need to compute the joint non-conformity score over all future times as in \Cref{eq:time_series_nonconf_score}.
\begin{gather}
    \bar{\mathcal{R}}_\tau :=\left\{ R_{S_c}^\tau(Z_\tau) \middle| Z_\tau \in D_{cal,1} \right\} \notag\\
    \beta_\tau := 1/\left(\text{Quantile}(\bar{\mathcal{R}}_\tau, 1-\delta) - \min(\bar{\mathcal{R}}_\tau)\right)
    \label{eq:time_normalizing_constant} \\
    R_{S_c}(Z) := \max_{\tau\in \{t+1, \ldots, T\}} \beta_\tau R_{S_c}^\tau(Z_\tau) \label{eq:time_series_nonconf_score}
\end{gather}
Note here that we take the maximum, as inspired by our prior work \cite{cleaveland2023conformal}, to obtain valid coverage over all future times. We can now apply conformal prediction to $R_{S_c}$ in the same way as in Theorem \ref{thm:cp} to obtain valid prediction regions for time series.
\section{Simulations}
\label{sec:exps}
The toolbox and all experiments below are available \href{https://github.com/nandantumu/conformal_region_designer}{on Github}.
We evaluate our approach using case studies on simulations of an F16 fighter jet performing ground avoidance maneuvers and a vehicle trajectory prediction scenario. Our method can be applied in two simple function calls after initialization.
\begin{minted}{python}
pcr = ConformityOptimizer("kde", "meanshift", "convexhull", 0.90)
pcr.fit(Z_cal_one)
pcr.conformalize(Z_cal_two)
\end{minted}


\subsection{F16}
\label{sec:f16}

In this case study, we analyze an F16 fighter jet performing ground avoidance maneuvers using the open source simulator from \cite{Heidlauf2018}. We use an LSTM to predict the altitude and pitch angle of the F16 up to $2.5$ seconds into the future at a rate of $10$ Hz ($25$ predictions in total) with the altitude and pitch from the previous $2.5$ seconds as input. The LSTM architecture consists of two layers of width $25$ and a final linear layer. We used $1500$ trajectories to train the network. 

\begin{figure}[h!]
    \centering
    \includegraphics[width=\linewidth]{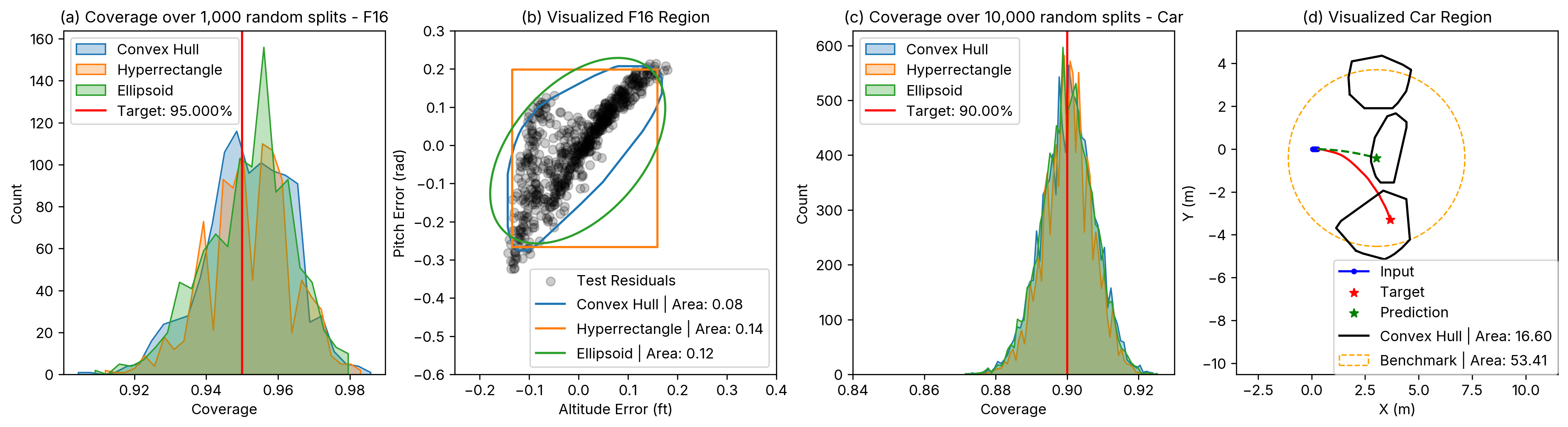}
    \caption{(a) Shows the coverage rates over 1,000 random splits of $D_{cal,2}$ and $D_{val}$ for the F16 example. (b) Shows fit conformal regions for the F16 example. (c) Shows the coverage rates over 10,000 random splits of $D_{cal,2}$ and $D_{val}$ for the car example. (d) Shows an example of the prediction regions shown on an actual prediction of the trajectories.}
    \label{fig:multi-coverage-plot}
\end{figure}

For calibration and validation, we collect a dataset of 1,900 trajectories all with length $5$ seconds. We randomly select 627 trajectories for $D_{cal,1}$, 627 for $D_{cal,2}$, and 646 for $D_{val}$. To account for the spread of the data, the bandwidth estimate was adjusted by a factor of $0.2$. Shapes were fit using the procedure described above, using a target coverage of $1 - \delta = 0.9$. The density estimation took $0.2949\text{s}$ on average and the clustering took $7.4222\text{s}$ on average. The average shape fitting times were $1.19\text{s}$ for the ellipse, $0.0030\text{s}$ for the convex hull, and $0.00086\text{s}$ for the hyperrectangle. Plots of the computed regions are shown in \Cref{fig:multi-coverage-plot}(b) and coverage over 1000 random splits of $D_{cal,2}$ and $D_{val}$ are shown in \Cref{fig:multi-coverage-plot}(a). The L2 norm benchmark region has a volume of $0.205$. Our regions provide a $31.7-60.9\%$ decrease in the area of the region, depending on the shape template used. In this example, differing units in each dimension are better compensated for in our approach.

\subsection{Vehicle Trajectory Prediction}
\label{sec:f1_tenth}
In this example application, we apply our method to the prediction of a vehicle's trajectory. The vehicle is governed according to kinematic dynamics, which are given by \Cref{eq:kinematic_dynamics}. The vehicle state contains its 2D position $x,y$, yaw $\theta$, and velocity $v$. The control inputs are the acceleration $a$ and steering angle $\omega$. For simplicity, we assume no acceleration commands (so $a=0$). 
\begin{equation} \label{eq:kinematic_dynamics}
    \dot{x} = v \cos(\theta),\quad 
    \dot{y} = v \sin(\theta),\quad 
    \dot{\theta} = v \tan(\omega)/L,\quad 
    \dot{v} = a;\quad
    s = [x,y,\theta,v],\quad
    u = [\omega,a]
\end{equation}


We use a physics-based, Constant Turn Rate and Velocity (CTRV) method to predict the trajectories of the car. The predictor takes as input the previous $0.5$ seconds of the state of the car (sampled at $10$Hz). It then estimates $\dot{\theta}$ by computing the average rate of change of $\theta$ over the inputs, and uses this estimate along with the current state of the car to predict the future position of the car up to $5$ seconds into the future at a rate of $10$Hz ($50$ predictions total) using \eqref{eq:kinematic_dynamics}.


The predictor is evaluated on a scenario which represents an intersection. The vehicle proceeds straight for $0.5$ seconds, then either proceeds forward, turns left, or turns right for $5$ seconds, all with equal probability. The predictor makes its predictions at the end of the $0.5$ straight period. $10000$ samples were generated, and split, with $3333$ samples in each $D_{cal,1}$ and $D_{cal,2}$, and $3334$ in the test set. 

First, we fit shapes for just the last timestep of the scenario, $5$ seconds into the prediction window, using the procedure described above with a target coverage of $1 - \delta = 0.9$. To account for the spread of the data, the bandwidth estimation was adjusted by a factor of $0.2$. We evaluated our approach on $10000$ random splits of the data in $D_{cal,1}$ and $D_{test}$. Computing the density estimate took $0.832\text{s}$ on average and the clusters took $0.856\text{s}$ on average. Fitting the shape templates took an average of $0.002\text{s}$ for the Convex Hull and Hyperrectangle and $3.760\text{s}$ for the Ellipse. The online portion, evaluating region membership, took $0.0029\text{s}$ on average for $3334$ points for all shapes. For each of the shape templates, we show that the mean coverage is close to our target coverage of $90\%$ in \Cref{fig:multi-coverage-plot}(c). The figure is shown in \Cref{fig:multi-coverage-plot}(d), and can be compared to the baseline circular region. Our method provides a $68.9\%$ improvement in the prediction region area while still providing the desired coverage. This figure also showcases the multi-modal capabilities of our approach, where each of the three behaviors has its own shape.

\begin{figure}[h]
    \centering
    \includegraphics[width=\linewidth]{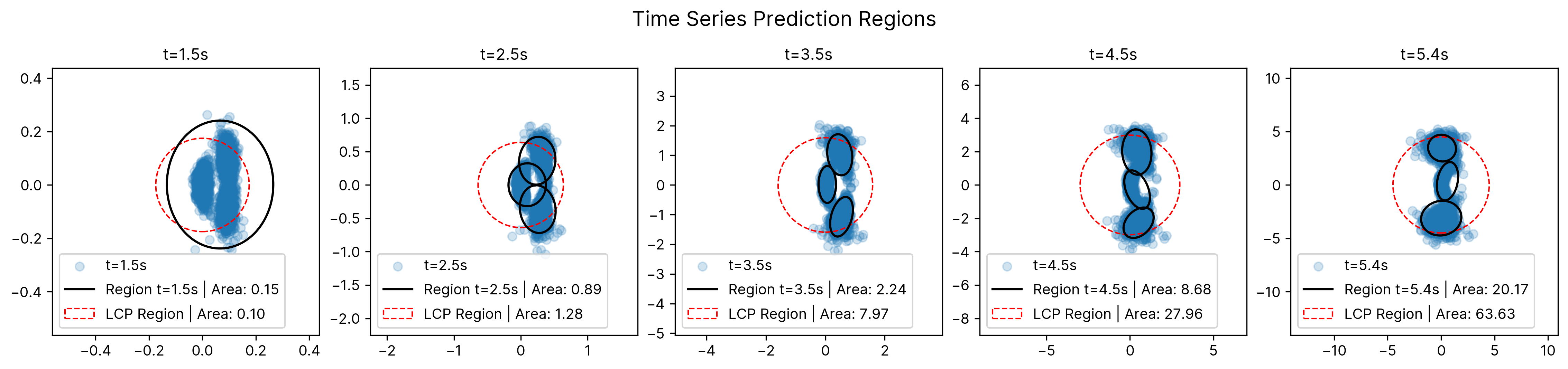}
    \caption{This figure shows the size of conformal prediction regions created for a time series prediction of vehicle motion over fifty timesteps. We generate a prediction region for only the timesteps shown, using the method in \cite{cleaveland2023conformal}, labeled the LCP region. We also generate conformal prediction regions using our method, which are shown in black. Each figure includes the area of the regions shown, and all methods achieve the desired coverage.}
    \label{fig:time-series}
\end{figure}

Finally, we computed regions over multiple timesteps. \Cref{fig:time-series} shows the size and shape of the regions designed to achieve $90\%$ coverage over 5 timesteps. This prediction region takes $28\text{s}$ to compute, and $0.008\text{s}$ to compute region membership. The total volume of the prediction region is $68.17\%$ smaller than the benchmark approach from \cite{cleaveland2023conformal}.

\section{Conclusion}
\label{sec:conclusion}



In this paper, we have presented a method for generating practical, multi-modal conformal prediction regions. Our approach uses an extra calibration dataset to find parameters of shape template functions over clusters of the calibration data. These shape template functions then get converted into a non-conformity score function, which we can use alongside standard inductive conformal prediction to get valid prediction regions. We demonstrate the approach on case studies of F16 fighter jets and autonomous vehicles, showing an up to 68\% reduction in prediction region area.
\clearpage

\section*{Acknowledgements}
This work was generously supported by NSF award SLES-2331880. Renukanandan Tumu was supported by NSF GRFP award DGE-2236662.

\bibliography{literature}

\begin{thebibliography}{41}
\providecommand{\natexlab}[1]{#1}
\providecommand{\url}[1]{\texttt{#1}}
\expandafter\ifx\csname urlstyle\endcsname\relax
  \providecommand{\doi}[1]{doi: #1}\else
  \providecommand{\doi}{doi: \begingroup \urlstyle{rm}\Url}\fi

\bibitem[Angelopoulos et~al.(2023)Angelopoulos, Bates,
  et~al.]{angelopoulos2021gentle}
Anastasios~N Angelopoulos, Stephen Bates, et~al.
\newblock Conformal prediction: A gentle introduction.
\newblock \emph{Foundations and Trends{\textregistered} in Machine Learning},
  16\penalty0 (4):\penalty0 494--591, 2023.

\bibitem[Angelopoulos et~al.(2024)Angelopoulos, Bates, Fisch, Lei, and
  Schuster]{angelopoulos2022conformal}
Anastasios~Nikolas Angelopoulos, Stephen Bates, Adam Fisch, Lihua Lei, and Tal
  Schuster.
\newblock Conformal risk control.
\newblock In \emph{The Twelfth International Conference on Learning
  Representations}, 2024.
\newblock URL \url{https://openreview.net/forum?id=33XGfHLtZg}.

\bibitem[Bai et~al.(2022)Bai, Mei, Wang, Zhou, and Xiong]{bai2022efficient}
Yu~Bai, Song Mei, Huan Wang, Yingbo Zhou, and Caiming Xiong.
\newblock Efficient and differentiable conformal prediction with general
  function classes.
\newblock In \emph{International Conference on Learning Representations}, 2022.
\newblock URL \url{https://openreview.net/forum?id=Ht85_jyihxp}.

\bibitem[Barber et~al.(1996)Barber, Dobkin, and
  Huhdanpaa]{barber_quickhull_1996}
C.~Bradford Barber, David~P. Dobkin, and Hannu Huhdanpaa.
\newblock The quickhull algorithm for convex hulls.
\newblock \emph{ACM Transactions on Mathematical Software}, 22\penalty0
  (4):\penalty0 469--483, December 1996.
\newblock ISSN 0098-3500.
\newblock \doi{10.1145/235815.235821}.

\bibitem[Cleaveland et~al.(2024)Cleaveland, Lee, Pappas, and
  Lindemann]{cleaveland2023conformal}
Matthew Cleaveland, Insup Lee, George~J. Pappas, and Lars Lindemann.
\newblock Conformal prediction regions for time series using linear
  complementarity programming.
\newblock \emph{Proceedings of the AAAI Conference on Artificial Intelligence},
  38\penalty0 (19):\penalty0 20984--20992, Mar. 2024.
\newblock \doi{10.1609/aaai.v38i19.30089}.
\newblock URL \url{https://ojs.aaai.org/index.php/AAAI/article/view/30089}.

\bibitem[Comaniciu and Meer(2002)]{comaniciu_mean_2002}
D.~Comaniciu and P.~Meer.
\newblock Mean shift: a robust approach toward feature space analysis.
\newblock \emph{IEEE Transactions on Pattern Analysis and Machine
  Intelligence}, 24\penalty0 (5):\penalty0 603--619, May 2002.
\newblock ISSN 1939-3539.
\newblock \doi{10.1109/34.1000236}.
\newblock Conference Name: IEEE Transactions on Pattern Analysis and Machine
  Intelligence.

\bibitem[Dixit et~al.(2023)Dixit, Lindemann, Wei, Cleaveland, Pappas, and
  Burdick]{dixit2023adaptive}
Anushri Dixit, Lars Lindemann, Skylar~X Wei, Matthew Cleaveland, George~J
  Pappas, and Joel~W Burdick.
\newblock Adaptive conformal prediction for motion planning among dynamic
  agents.
\newblock In \emph{Learning for Dynamics and Control Conference}, pages
  300--314. PMLR, 2023.

\bibitem[Fannjiang et~al.(2022)Fannjiang, Bates, Angelopoulos, Listgarten, and
  Jordan]{fannjiang2022conformal}
Clara Fannjiang, Stephen Bates, Anastasios~N Angelopoulos, Jennifer Listgarten,
  and Michael~I Jordan.
\newblock Conformal prediction under feedback covariate shift for biomolecular
  design.
\newblock \emph{Proceedings of the National Academy of Sciences}, 119\penalty0
  (43):\penalty0 e2204569119, 2022.

\bibitem[Feldman et~al.(2023)Feldman, Bates, and
  Romano]{feldman_calibrated_2022}
Shai Feldman, Stephen Bates, and Yaniv Romano.
\newblock Calibrated multiple-output quantile regression with representation
  learning.
\newblock \emph{Journal of Machine Learning Research}, 24\penalty0
  (24):\penalty0 1--48, 2023.

\bibitem[Han et~al.(2022)Han, Tang, Ghosh, and Liu]{han2022split}
Xing Han, Ziyang Tang, Joydeep Ghosh, and Qiang Liu.
\newblock Split localized conformal prediction.
\newblock \emph{arXiv preprint arXiv:2206.13092}, 2022.

\bibitem[Hansen et~al.(2003)Hansen, Müller, and
  Koumoutsakos]{hansen_reducing_2003}
Nikolaus Hansen, Sibylle~D. Müller, and Petros Koumoutsakos.
\newblock Reducing the {Time} {Complexity} of the {Derandomized} {Evolution}
  {Strategy} with {Covariance} {Matrix} {Adaptation} ({CMA}-{ES}).
\newblock \emph{Evolutionary Computation}, 11\penalty0 (1):\penalty0 1--18,
  March 2003.
\newblock ISSN 1063-6560.
\newblock \doi{10.1162/106365603321828970}.

\bibitem[Hansen et~al.(2023)Hansen, {Yoshihikoueno}, {ARF1}, Kadlecová,
  Nozawa, Rolshoven, Chan, {Youhei Akimoto}, {Brieglhostis}, and
  Brockhoff]{hansen_cma-espycma_2023}
Nikolaus Hansen, {Yoshihikoueno}, {ARF1}, Gabriela Kadlecová, Kento Nozawa,
  Luca Rolshoven, Matthew Chan, {Youhei Akimoto}, {Brieglhostis}, and Dimo
  Brockhoff.
\newblock {CMA}-{ES}/pycma: r3.3.0, January 2023.

\bibitem[Hashemi et~al.(2023)Hashemi, Qin, Lindemann, and
  Deshmukh]{hashemi2023data}
Navid Hashemi, Xin Qin, Lars Lindemann, and Jyotirmoy~V. Deshmukh.
\newblock Data-driven reachability analysis of stochastic dynamical systems
  with conformal inference.
\newblock In \emph{2023 62nd IEEE Conference on Decision and Control (CDC)},
  pages 3102--3109, 2023.
\newblock \doi{10.1109/CDC49753.2023.10384213}.

\bibitem[Heidlauf et~al.(2018)Heidlauf, Collins, Bolender, and
  Bak]{Heidlauf2018}
Peter Heidlauf, Alexander Collins, Michael Bolender, and Stanley Bak.
\newblock Reliable prediction intervals with directly optimized inductive
  conformal regression for deep learning.
\newblock \emph{5th International Workshop on Applied Verification for
  Continuous and Hybrid Systems (ARCH 2018)}, 2018.

\bibitem[Izbicki et~al.(2022)Izbicki, Shimizu, and Stern]{Izbicki2022}
Rafael Izbicki, Gilson Shimizu, and Rafael~B. Stern.
\newblock Cd-split and hpd-split: Efficient conformal regions in high
  dimensions.
\newblock \emph{J. Mach. Learn. Res.}, 23\penalty0 (1), jan 2022.
\newblock ISSN 1532-4435.

\bibitem[Kaur et~al.(2022)Kaur, Jha, Roy, Park, Dobriban, Sokolsky, and
  Lee]{Kaur2022}
Ramneet Kaur, Susmit Jha, Anirban Roy, Sangdon Park, Edgar Dobriban, Oleg
  Sokolsky, and Insup Lee.
\newblock idecode: In-distribution equivariance for conformal
  out-of-distribution detection.
\newblock \emph{Proceedings of the AAAI Conference on Artificial Intelligence},
  36\penalty0 (7):\penalty0 7104--7114, Jun. 2022.
\newblock \doi{10.1609/aaai.v36i7.20670}.

\bibitem[Kaur et~al.(2023{\natexlab{a}})Kaur, Sridhar, Park, Yang, Jha, Roy,
  Sokolsky, and Lee]{kaur2022codit}
Ramneet Kaur, Kaustubh Sridhar, Sangdon Park, Yahan Yang, Susmit Jha, Anirban
  Roy, Oleg Sokolsky, and Insup Lee.
\newblock Codit: Conformal out-of-distribution detection in time-series data
  for cyber-physical systems.
\newblock In \emph{Proceedings of the ACM/IEEE 14th International Conference on
  Cyber-Physical Systems (with CPS-IoT Week 2023)}, ICCPS '23, page 120–131,
  New York, NY, USA, 2023{\natexlab{a}}. Association for Computing Machinery.
\newblock ISBN 9798400700361.
\newblock \doi{10.1145/3576841.3585931}.
\newblock URL \url{https://doi.org/10.1145/3576841.3585931}.

\bibitem[Kaur et~al.(2023{\natexlab{b}})Kaur, Sridhar, Park, Yang, Jha, Roy,
  Sokolsky, and Lee]{kaur2023codit}
Ramneet Kaur, Kaustubh Sridhar, Sangdon Park, Yahan Yang, Susmit Jha, Anirban
  Roy, Oleg Sokolsky, and Insup Lee.
\newblock Codit: Conformal out-of-distribution detection in time-series data
  for cyber-physical systems.
\newblock In \emph{Proceedings of the ACM/IEEE 14th International Conference on
  Cyber-Physical Systems (with CPS-IoT Week 2023)}, pages 120--131,
  2023{\natexlab{b}}.

\bibitem[Lei and Wasserman(2014)]{lei2014distribution}
Jing Lei and Larry Wasserman.
\newblock Distribution-free prediction bands for non-parametric regression.
\newblock \emph{Journal of the Royal Statistical Society Series B: Statistical
  Methodology}, 76\penalty0 (1):\penalty0 71--96, 2014.

\bibitem[Lei et~al.(2011)Lei, Robins, and Wasserman]{lei2011efficient}
Jing Lei, James Robins, and Larry Wasserman.
\newblock Efficient nonparametric conformal prediction regions.
\newblock \emph{arXiv preprint arXiv:1111.1418}, 2011.

\bibitem[Lei et~al.(2013)Lei, Robins, and Wasserman]{lei2013distribution}
Jing Lei, James Robins, and Larry Wasserman.
\newblock Distribution-free prediction sets.
\newblock \emph{Journal of the American Statistical Association}, 108\penalty0
  (501):\penalty0 278--287, 2013.

\bibitem[Lindemann et~al.(2023{\natexlab{a}})Lindemann, Cleaveland, Shim, and
  Pappas]{lindemann2022safe}
Lars Lindemann, Matthew Cleaveland, Gihyun Shim, and George~J. Pappas.
\newblock Safe planning in dynamic environments using conformal prediction.
\newblock \emph{IEEE Robotics and Automation Letters}, 8\penalty0 (8):\penalty0
  5116--5123, 2023{\natexlab{a}}.
\newblock \doi{10.1109/LRA.2023.3292071}.

\bibitem[Lindemann et~al.(2023{\natexlab{b}})Lindemann, Qin, Deshmukh, and
  Pappas]{lindemann2022conformal}
Lars Lindemann, Xin Qin, Jyotirmoy~V. Deshmukh, and George~J. Pappas.
\newblock Conformal prediction for stl runtime verification.
\newblock In \emph{Proceedings of the ACM/IEEE 14th International Conference on
  Cyber-Physical Systems (with CPS-IoT Week 2023)}, ICCPS '23, page 142–153,
  2023{\natexlab{b}}.
\newblock ISBN 9798400700361.
\newblock \doi{10.1145/3576841.3585927}.

\bibitem[Luo et~al.(2022)Luo, Zhao, Kuck, Ivanovic, Savarese, Schmerling, and
  Pavone]{luo2022sample}
Rachel Luo, Shengjia Zhao, Jonathan Kuck, Boris Ivanovic, Silvio Savarese,
  Edward Schmerling, and Marco Pavone.
\newblock Sample-efficient safety assurances using conformal prediction.
\newblock In \emph{Algorithmic Foundations of Robotics XV: Proceedings of the
  Fifteenth Workshop on the Algorithmic Foundations of Robotics}, pages
  149--169. Springer, 2022.

\bibitem[Papadopoulos(2008)]{papadopoulos2008inductive}
Harris Papadopoulos.
\newblock Inductive conformal prediction: Theory and application to neural
  networks.
\newblock In \emph{Tools in artificial intelligence}. Citeseer, 2008.

\bibitem[Parente et~al.(2023)Parente, Darabi, Stutts, Tulabandhula, and
  Trivedi]{parente2023conformalized}
Domenico Parente, Nastaran Darabi, Alex~C Stutts, Theja Tulabandhula, and
  Amit~Ranjan Trivedi.
\newblock Conformalized multimodal uncertainty regression and reasoning.
\newblock \emph{arXiv preprint arXiv:2309.11018}, 2023.

\bibitem[Parzen(1962)]{parzen_estimation_1962}
Emanuel Parzen.
\newblock On {Estimation} of a {Probability} {Density} {Function} and {Mode}.
\newblock \emph{The Annals of Mathematical Statistics}, 33\penalty0
  (3):\penalty0 1065--1076, 1962.
\newblock ISSN 0003-4851.
\newblock Publisher: Institute of Mathematical Statistics.

\bibitem[Pedregosa et~al.(2011)Pedregosa, Varoquaux, Gramfort, Michel, Thirion,
  Grisel, Blondel, Prettenhofer, Weiss, Dubourg, Vanderplas, Passos,
  Cournapeau, Brucher, Perrot, and Duchesnay]{scikit-learn}
F.~Pedregosa, G.~Varoquaux, A.~Gramfort, V.~Michel, B.~Thirion, O.~Grisel,
  M.~Blondel, P.~Prettenhofer, R.~Weiss, V.~Dubourg, J.~Vanderplas, A.~Passos,
  D.~Cournapeau, M.~Brucher, M.~Perrot, and E.~Duchesnay.
\newblock Scikit-learn: Machine learning in {P}ython.
\newblock \emph{Journal of Machine Learning Research}, 12:\penalty0 2825--2830,
  2011.

\bibitem[Romano et~al.(2019)Romano, Patterson, and
  Candes]{romano2019conformalized}
Yaniv Romano, Evan Patterson, and Emmanuel Candes.
\newblock Conformalized quantile regression.
\newblock \emph{Advances in neural information processing systems}, 32, 2019.

\bibitem[Rosenblatt(1956)]{rosenblatt_remarks_1956}
Murray Rosenblatt.
\newblock Remarks on {Some} {Nonparametric} {Estimates} of a {Density}
  {Function}.
\newblock \emph{The Annals of Mathematical Statistics}, 27\penalty0
  (3):\penalty0 832--837, September 1956.
\newblock ISSN 0003-4851, 2168-8990.
\newblock \doi{10.1214/aoms/1177728190}.
\newblock Publisher: Institute of Mathematical Statistics.

\bibitem[Shafer and Vovk(2008)]{shafer2008tutorial}
Glenn Shafer and Vladimir Vovk.
\newblock A tutorial on conformal prediction.
\newblock \emph{Journal of Machine Learning Research}, 9\penalty0 (3), 2008.

\bibitem[Silverman(1986)]{silverman_density_1986}
Bernard~W. Silverman.
\newblock \emph{Density {Estimation} for {Statistics} and {Data} {Analysis}}.
\newblock CRC Press, April 1986.
\newblock ISBN 978-0-412-24620-3.

\bibitem[Smith et~al.(2014)Smith, Nouretdinov, Craddock, Offer, and
  Gammerman]{Smith2014ConformalKDE}
James Smith, Ilia Nouretdinov, Rachel Craddock, Charles Offer, and Alexander
  Gammerman.
\newblock Anomaly detection of trajectories with kernel density estimation by
  conformal prediction.
\newblock In Lazaros Iliadis, Ilias Maglogiannis, Harris Papadopoulos, Spyros
  Sioutas, and Christos Makris, editors, \emph{Artificial Intelligence
  Applications and Innovations}, pages 271--280, Berlin, Heidelberg, 2014.
  Springer Berlin Heidelberg.

\bibitem[Stutz et~al.(2022)Stutz, Dvijotham, Cemgil, and Doucet]{stutz2022ICLR}
David Stutz, Krishnamurthy~Dj Dvijotham, Ali~Taylan Cemgil, and Arnaud Doucet.
\newblock Learning optimal conformal classifiers.
\newblock In \emph{International Conference on Learning Representations}, 2022.

\bibitem[Tibshirani et~al.(2019)Tibshirani, Foygel~Barber, Candes, and
  Ramdas]{tibshirani2019conformal}
Ryan~J Tibshirani, Rina Foygel~Barber, Emmanuel Candes, and Aaditya Ramdas.
\newblock Conformal prediction under covariate shift.
\newblock \emph{Advances in neural information processing systems}, 32, 2019.

\bibitem[Tumu et~al.(2023)Tumu, Lindemann, Nghiem, and
  Mangharam]{tumu2023physics}
Renukanandan Tumu, Lars Lindemann, Truong Nghiem, and Rahul Mangharam.
\newblock Physics {Constrained} {Motion} {Prediction} with {Uncertainty}
  {Quantification}.
\newblock In \emph{2023 {IEEE} {Intelligent} {Vehicles} {Symposium} ({IV})},
  pages 1--8, Anchorage, AK, USA, June 2023. IEEE.
\newblock \doi{10.1109/IV55152.2023.10186812}.

\bibitem[Vovk(2012)]{vovk2012conditional}
Vladimir Vovk.
\newblock Conditional validity of inductive conformal predictors.
\newblock In \emph{Asian conference on machine learning}, pages 475--490. PMLR,
  2012.

\bibitem[Vovk et~al.(2005)Vovk, Gammerman, and Shafer]{vovk2005algorithmic}
Vladimir Vovk, Alexander Gammerman, and Glenn Shafer.
\newblock \emph{Algorithmic learning in a random world}.
\newblock Springer Science \& Business Media, 2005.

\bibitem[Wang et~al.(2023{\natexlab{a}})Wang, Gao, Yin, Zhou, and
  Blei]{wang2022probabilistic}
Zhendong Wang, Ruijiang Gao, Mingzhang Yin, Mingyuan Zhou, and David Blei.
\newblock Probabilistic conformal prediction using conditional random samples.
\newblock In \emph{Proceedings of The 26th International Conference on
  Artificial Intelligence and Statistics}, pages 8814--8836, 25--27 Apr
  2023{\natexlab{a}}.

\bibitem[Wang et~al.(2023{\natexlab{b}})Wang, Gao, Yin, Zhou, and
  Blei]{wang_probabilistic_2022}
Zhendong Wang, Ruijiang Gao, Mingzhang Yin, Mingyuan Zhou, and David Blei.
\newblock Probabilistic conformal prediction using conditional random samples.
\newblock In Francisco Ruiz, Jennifer Dy, and Jan-Willem van~de Meent, editors,
  \emph{Proceedings of The 26th International Conference on Artificial
  Intelligence and Statistics}, volume 206 of \emph{Proceedings of Machine
  Learning Research}, pages 8814--8836. PMLR, 25--27 Apr 2023{\natexlab{b}}.
\newblock URL \url{https://proceedings.mlr.press/v206/wang23n.html}.

\bibitem[Zecchin et~al.(2023)Zecchin, Park, and Simeone]{zecchin2023forking}
Matteo Zecchin, Sangwoo Park, and Osvaldo Simeone.
\newblock Forking uncertainties: Reliable prediction and model predictive
  control with sequence models via conformal risk control.
\newblock \emph{arXiv preprint arXiv:2310.10299}, 2023.

\end{thebibliography}


\addtolength{\textheight}{-12cm}   

\end{document}